%% file: main.tex
\documentclass[natbib,smallcondensed]{svjour3}
\usepackage[utf8]{inputenc}

\usepackage{subcaption}
\usepackage{graphicx}

\usepackage{mathtools}
\usepackage{amsfonts}       
\usepackage{bbm}
\usepackage{longtable}
\usepackage{multirow}
\usepackage{textcomp}
\newcommand{\numberset}{\mathbb} %

\newcommand{\E}{\numberset{E}}

\newcommand{\Var}{\mathrm{Var}}

\DeclareMathOperator*{\argmin}{arg\,min}
\usepackage{tikz}
\usepackage{ctable} 
\usepackage{adjustbox}
\usepackage{bbm}
\usepackage[normalem]{ulem}

\usepackage[hidelinks]{hyperref}
\usepackage{mathtools}

\newcommand{\changed}[1]{{#1}}

\newcommand{\rr}[1]{\text{\textregistered}}

\usepackage{scalerel,stackengine}
\stackMath
\newcommand\reallywidehat[1]{%
\savestack{\tmpbox}{\stretchto{%
  \scaleto{%
    \scalerel*[\widthof{\ensuremath{#1}}]{\kern-.6pt\bigwedge\kern-.6pt}%
    {\rule[-\textheight/2]{1ex}{\textheight}}
  }{\textheight}%
}{0.5ex}}%
\stackon[1pt]{#1}{\tmpbox}%
}

\usepackage{glossaries}
\newacronym{svdd}{SVDD}{Support Vector Data Descriptor}
\newacronym{svm}{SVM}{Support Vector Machine}
\newacronym{ocsvm}{OCSVM}{One-Class Support Vector Machine}
\newacronym{lof}{LOF}{Local Outlier Factor}
\newacronym{if}{IF}{Isolation Forest}
\newacronym{gmm}{GMM}{Gaussian Mixture Model}
\newacronym{hmm}{HMM}{Hidden Markov Model}
\newacronym{vae}{VAE}{Variational Autoencoders}
\newacronym{knn}{k-NN}{k-Nearest Neighbors}
\newacronym{knno}{k-NNO}{k-Nearest Neighbors Outlier}
\newacronym{som}{SOM}{Self-Organizing Map}
\newacronym{rbfn}{RBFN}{Radial Basis Function Network}
\newacronym{lvq}{LVQ}{Learning Vector Quantization}
\newacronym{smo}{SMO}{Sequential Minimal Optimization}
\newacronym{sgd}{SGD}{Stochastic Gradient Descent}
\newacronym{gpd}{GPD}{Generalized Probabilistic Descent}

\newacronym{roc}{ROC}{Receiver Operating Characteristic}
\newacronym{pr}{PR}{Precision-Recall}
\newacronym{tpr}{TPR}{True Postive Rate}
\newacronym{fpr}{FPR}{False Postive Rate}
\newacronym{arc}{ARC}{Accuracy-Reject Curve}
\newacronym{auc}{AUC}{Area Under the ROC Curve}
\newacronym{aurc}{AURC}{Area Under the ARC Curve}

\begin{document}
\title{Machine Learning with a Reject Option: A survey}
\author{Hendrickx Kilian* \and Perini Lorenzo* \and Van der Plas Dries \and Meert Wannes \and Davis Jesse}
\authorrunning{Hendrickx K., Perini L., et al}

\institute{* These authors contributed equally to this work. \\ 
	\and
K. Hendrickx \at
              Siemens Digital Industries Software, Leuven, Belgium \\
              KU Leuven, Leuven, Belgium \\
              \email{kilian.hendrickx@cs.kuleuven.be}           
          \and
          L. Perini \at
              KU Leuven, Leuven, Belgium \\
              \email{lorenzo.perini@cs.kuleuven.be}           
          \and
          D. Van der Plas \at
              OSG bv, Natus Medical, Kontich, Belgium \\
              KU Leuven, Leuven, Belgium \\
              University of Antwerp, Antwerp, Belgium \\
              \email{dries.vanderplas@cs.kuleuven.be}           
          \and
          W. Meert \at
              KU Leuven; Leuven.AI, Leuven, Belgium \\
              \email{wannes.meert@cs.kuleuven.be}           
          \and
          J. Davis \at
              KU Leuven; Leuven.AI, Leuven, Belgium \\
              \email{jesse.davis@cs.kuleuven.be}           
              \and
              Corresponding author: Kilian Hendrickx
}

\date{Received: 23 July 2021 / Revised: 27 July 2023 \& 15 February 2024}

\maketitle

\begin{abstract}
Machine learning models always make a prediction, even when it is likely to be inaccurate. This behavior should be avoided in many decision support applications, where mistakes can have severe consequences. Albeit already studied in 1970, machine learning with rejection recently gained interest. This machine learning subfield enables machine learning models to abstain from making a prediction when likely to make a mistake.

This survey aims to provide an overview on machine learning with rejection. We introduce the conditions leading to two types of rejection, ambiguity and novelty rejection, which we carefully formalize. Moreover, we review and categorize strategies to evaluate a model's predictive and rejective quality. Additionally, we define the existing architectures for models with rejection and describe the standard techniques for learning such models. Finally, we provide examples of relevant application domains and show how machine learning with rejection relates to other machine learning research areas. 


\end{abstract}

\tableofcontents
\newpage
\input{a_introduction}

\input{b_preliminaries}
\input{c_metrics}
\input{d_SeparatedRejector}

\input{e_DependentRejector}

\input{f_IntegratedRejector}
\input{g_CombineRejectors}

\input{h_applications}
\input{i_related_areas}
\input{j_conclusions}

\section*{Declarations}

\subsection*{Funding}
Kilian Hendrickx and Dries Van der Plas received funding from VLAIO (Flemish Innovation \& Entrepreneurship) through the Baekeland PhD mandates [HBC.2017.0226] (KH) and [HBC.2019.2615] (DV). 
Lorenzo Perini received funding from FWO-Vlaanderen, aspirant grant 1166222N.
Jesse Davis is partially supported by the KU Leuven research funds [C14/17/070]. 
Lorenzo Perini, Jesse Davis and Wannes Meert received funding from the Flemish Government under the ``Onderzoeksprogramma Artifici\"{e}le Intelligentie (AI) Vlaanderen'' programme.

\subsection*{Conflict of interest}
The authors declare that they have no conflict of interest.

\subsection*{Ethics approval}
Not applicable

\subsection*{Consent to participate}
Not applicable

\subsection*{Availability of data and material}
Not applicable

\subsection*{Code availability}
Not applicable

\subsection*{Authors' contributions}
Concept: JD, WM;
Literature Study: KH, LP, DVdP; 
Categorization: KH, LP, DVdP, WM, JD;
Writing - original draft preparation: KH, LP, DVdP;
Writing - review and editing: WM, JD;
Writing - revision: LP, WM, JD, DVdP, KH;
Funding acquisition: WM, JD;
Supervision: WM, JD.

\bibliographystyle{apalike}
\bibliography{references}

\end{document}

%% file: a_introduction.tex
\section{Introduction}
The canonical task in machine learning is to learn a predictive model that captures the relationship between a set of input variables and a target variable on the basis of training data. Machine-learned models are powerful because, after training, they offer the ability to make accurate predictions about future examples. Since this enables automating a number of tasks that are difficult and/or time-consuming, such models are ubiquitously deployed. 

However, their key functionality of always returning a prediction for a given novel input is also a drawback. While the model may produce accurate predictions in general, in certain circumstances this may not be the case. For instance, there could be certain regions of the feature space where the model struggles to differentiate among the different classes. Or the current test example could be highly dissimilar to the data used to train the model. In certain application domains, such as medical diagnostics~\citep{Kotropoulos2009} and engineering~\citep{Zou2011}, mispredictions can have serious consequences. Therefore, it would be beneficial for a model to be cautious in situations where it is uncertain about its predictions. The prediction task could be deferred to a human expert in these situations. 

One way to accomplish this is to use machine learning models with rejection. Such models assess their confidence in each prediction and have the option to abstain from making a prediction when they are likely to make a mistake. This ability to abstain from making a prediction has several benefits. First, by only making predictions when it is confident, it can result in improved performance for the retained examples~\citep{Pudil1992}. Second, avoiding mispredictions can increase a user's trust in the system~\citep{El-Yaniv2010}. Third, it can still result in time savings by only requiring human interventions to make decisions in a small number of cases. Fourth, avoiding strongly biased predictions helps build a more fair model~\citep{Lee2021,Ruggieri2023}.
This machine learning sub-field was already studied in 1970 by \citet{Chow1970a} and \citet{Hellman1970}. However, the proliferation of applications has resulted in renewed interest in this area.

This survey aims to provide an overview of the subfield of machine learning with rejection, which we structure around eight key research questions.
\begin{enumerate}
    \item[Q1.] How can we formalize the conditions for which a model should abstain from making a prediction?
    \item[Q2.] How can we evaluate the performance of a model with rejection?
    \item[Q3.] What architectures are possible for operationalizing (i.e., putting this into practice) the ability to abstain from making a prediction?
    \item[Q4.] How do we learn models with rejection?
    \item[Q5.] What are the main pros and cons of using a specific architecture?
    \item[Q6.] How can we combine multiple rejectors?
    \item[Q7.] Where does the need for machine learning with rejection methods arise in real-world applications?
    \item[Q8.] How does machine learning with rejection relate to other research areas?
\end{enumerate}
In addition to the individual contributions of addressing each of these research questions, our major contribution is that we identify the main characteristics of machine learning models with rejection, allowing us to structure the methods in this research field.
By providing an overview of the research field as well as deeper insights into the various techniques, we aid in further advance this research area, as well as its adaptation to real-world applications. 

The remainder of this paper is structured as follows. In Section~\ref{sec:preliminaries}, we formalize the setting in which machine learning with rejection operates and identify the two main motivations to abstain from making a prediction (Q1). Section~\ref{sec:metrics} introduces the means to evaluate the performance of models with rejection (Q2).
Sections~\ref{sec:separated_rejector},~\ref{sec:dependent_rejector}, and~\ref{sec:integrated_rejector} provide a structured overview of the actionable techniques to reject based on the relevant literature. In these sections, we focus on describing the architecture (Q3), the rejector's learning (Q4), and the key pros and cons (Q5). In Section~\ref{sec:combining_rejectors} we explore how to combine multiple rejectors to allow different types of rejection (Q6)
Section~\ref{sec:applications} discusses the main application fields (Q7), while Section~\ref{sec:related_research_areas} explores the relation of machine learning with rejection with other research areas (Q8). 
Finally, Section~\ref{sec:conclusion_and_future_direction} summarizes our conclusions and lists the main open research questions.

%% file: b_preliminaries.tex
\section{The learning with reject problem setting}\label{sec:preliminaries}

In the standard supervised setting, a learner has access to a training set $D = \{(x_1,y_1),\ldots, (x_n,y_n)\}$, where each $x_i$ is a $d$ dimensional vector and $y_i$ is the target. The training data is assumed to be independent and identically distributed (i.i.d.) according to some unknown probability measure $P$ (with density $p(X,Y)$).
More generally, we denote the feature space as $\mathcal{X}$ and the target space as $\mathcal{Y}$, which could be discrete $\mathcal{Y} = \{1, 2, \ldots, K\}$, continuous  $\mathcal{Y} = \mathbb{R}$, or even probabilistic $\mathcal{Y} = [0,1]$. 

The assumption is that there is an unknown, non-deterministic function $f\colon \mathcal{X} \to \mathcal{Y}$ that maps the examples to their target value. Given a hypothesis space $\mathcal{H}$ of functions $h \colon \mathcal{X} \to \mathcal{Y}$, the goal of a learner is to find a good approximation to $f$. Typically, this can be done by finding a model $h \in \mathcal{H}$ with a small expected risk $R$ which is usually approximated using the training data
\begin{equation}
    R(h) \coloneqq \int_{\mathcal{X} \times \mathcal{Y}} L(h(x),y) d P(x,y) \approx \sum_{i=1}^{n} \frac{L(h(x_i),y_i)}{n},
\end{equation}
where $L$ is a suitable loss function such as the squared or zero-one loss.


\subsection{Models with a reject option}
In learning with rejection, the output space of the model is extended to include a new value $\rr{}$~\citep{Cortes2016,pmlr-v80-cortes18a,Sousa2014a}. This new symbol means that the model abstains from making a prediction. \changed{When performing classification with rejection, which is also called cautious classification, this new output $\rr{}$ can be seen as an additional class~\citep{Ferri2004}.}

Conceptually, since $h$ only approximates the true underlying model, there are likely regions of $\mathcal{X}$ where $h$ systematically differs from $f$. 
Specifically, discrepancies between $h$ and $f$ can be due to inconsistent data (e.g., classes overlapping), insufficient data (e.g., unexplored regions), or even incorrect model assumptions (e.g., $h$ must be linear while $f$ is not).
Therefore, the goal of rejection is to determine such regions in order to abstain from making likely inaccurate predictions.

Formally, a model with rejection $m \colon \mathcal{X} \to \mathcal{Y} \cup \{\rr{}\}$ is represented by a pair $(h, r)$, where $h \colon \mathcal{X} \to \mathcal{Y}$ is the predictor and $r \colon \mathcal{R} \to \mathbb{R}$ is the rejector. Note that the rejector may use a variety of different inputs such as examples ($\mathcal{R} = \mathcal{X}$), confidence or probability values ($\mathcal{R} = [0,1]$), or even both ($\mathcal{R} = \mathcal{X} \times [0,1]$).
At prediction time, $m$ outputs the symbol $\rr{}$ and abstains from making predictions when the rejector $r$ determines that the predictor is at a heightened risk of making a misprediction and otherwise returns the predictor's output:
\begin{equation}\label{eq:general_modelwithreject_equation1}
m(x) = \begin{cases}
    \rr{} &\text{if the prediction is \emph{rejected}};\\
    h(x) &\text{if the prediction is \emph{accepted}}.
\end{cases} 
\end{equation}

\begin{figure}[th]
	\centering
	\includegraphics[width=0.8\linewidth]{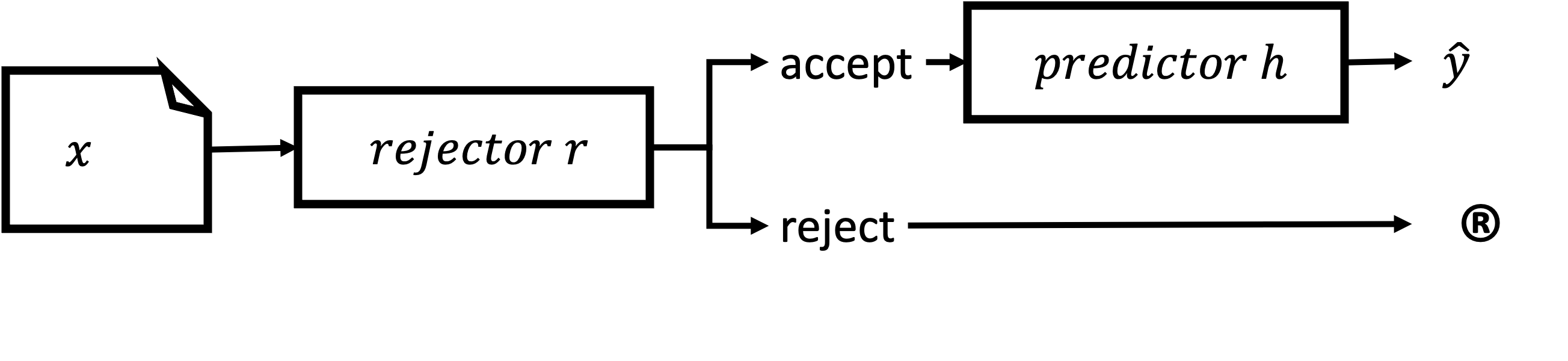}
	\caption{Data flow in a separated rejector, in which both the predictor and the rejector are stand-alone models. The rejector is a filter and only passes accepted examples to the predictor.}
	\label{fig:level_1_predict}
\end{figure}

\begin{figure}[th]
\centering
\includegraphics[width=0.8\linewidth]{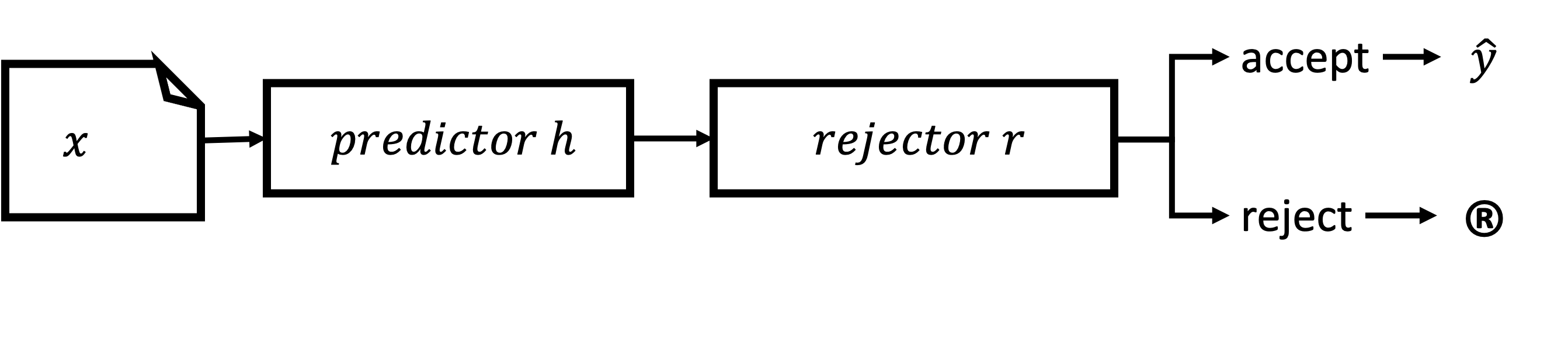}
\caption{Data flow in a dependent rejector. First, the predictor processes the example at hand. Next, the rejector assesses the confidence in the prediction based on the predictor's representation of the example.}
\label{fig:level_2_predict}
\end{figure}

\begin{figure}[th]
	\centering
	\includegraphics[width=0.8\linewidth]{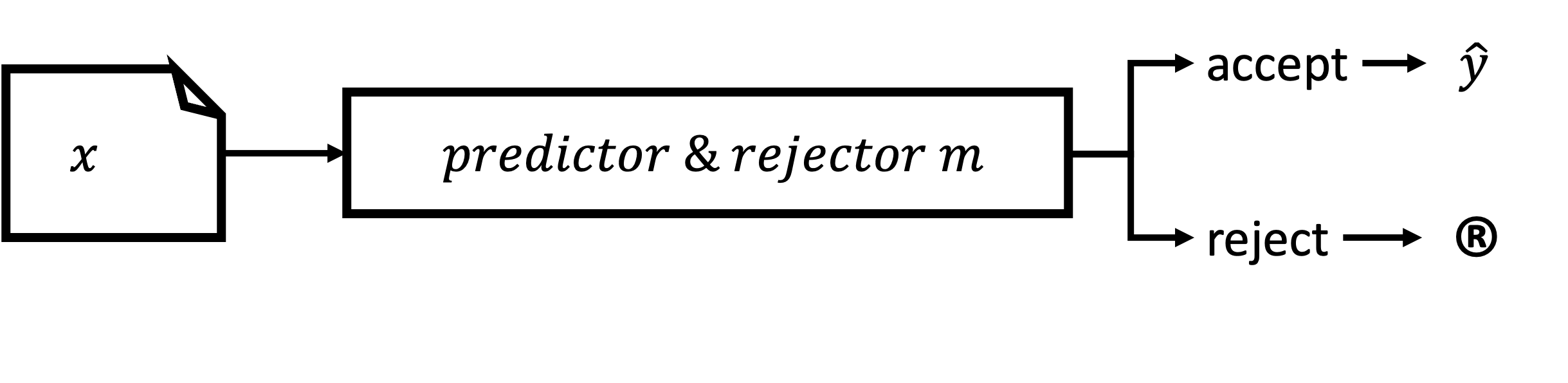}
	\caption{Data flow in an integrated rejector, in which the predictor and rejector are one model. This model embeds the reject and predict functions and directly outputs a prediction or a rejection.}
	\label{fig:level_3_predict}
\end{figure}

At prediction time, the key design decision for a model with rejection is how to structure the relationship between the predictor and rejector. Based on our analysis of the literature, we have identified three common architectural principles.
\begin{description}
\item[{\bf Separated rejector architecture.}] The rejector operates independently from the predictor. The most typical operationalization of this architecture lets the rejector serve as a filter that decides whether to pass a test example to the predictor.
Figure~\ref{fig:level_1_predict} shows the test time data flow of a separated rejector. 
\item[{\bf Dependent rejector architecture.}] Here, the rejector bases its decision on the output of the predictor. For instance, the dependent rejector can look at how close a prediction is to the predictor's decision boundary and abstain if it is too close. Figure~\ref{fig:level_2_predict} shows the data flow for the dependent rejector. 
\item[{\bf Integrated rejector architecture.}] This principle involves integrating the rejector and predictor into a single model $m$ by treating rejection as an additional class that can be returned by the model $m$. Thus, it is impossible to distinguish between the role of the predictor $h$ and the rejector $r$.
Figure~\ref{fig:level_3_predict} illustrates this scenario.
\end{description}

\subsection{Types of rejection}

\changed{At a high-level, a learned predictor can exhibit (high) uncertainty in its predictions for four reasons:}
\begin{enumerate}
    \item[\textbf{R1.}] \changed{There can be cases where a vector $x_i \in \mathcal{X}$ is associated with multiple values from the target space $\mathcal{Y}$. This can arise in situations such as when classes overlap in classification tasks.} 
    \item[\textbf{R2.}] \changed{Some instances in the training data are incorrect (e.g., the values of certain features were recorded incorrectly, labeling errors).}
    \item[\textbf{R3.}] \changed{$P(X,Y)$ might differ between the training phase and deployment (e.g., concept drift). Consequently, the training data is no longer representative.} 
    \item[\textbf{R4.}] \changed{Some examples $x$ could simply not be acquired due to their inherent rarity (anomalies, out-of-distribution).}
\end{enumerate}

\changed{\noindent Based on this intuition, two types of rejections can be performed:}
\begin{description}
    \item[\emph{\textbf{Ambiguity Rejection:}}] \changed{occurs if $x$ falls in a region where the target $y$ is ambiguous (\textbf{R1} and \textbf{R2}). This often occurs in regions that are close to the decision boundary in classification tasks;}
    \item[\emph{\textbf{Novelty Rejection:}}] \changed{occurs if $x$ falls in a region where there was little (or no) training data. Hence, the predictor may struggle to make accurate predictions because it did not see enough data to accurately model the relationship between $X$ and $Y$ (\textbf{R3} and \textbf{R4}).}
    
\end{description}

\subsubsection{Ambiguity Rejection}

\changed{\textit{Ambiguity rejection} allows a model to abstain from making a prediction for an example $x$ in regions where, despite having access to some training examples, the model $h$ fails to capture the correct relationship $f$ between $X$ and $Y$~\citep{Chow1958,Hellman1970,Fukunaga1972}. This can happen for two reasons.} 

\changed{First, the observed relationship between $X$ and $Y$ is not deterministic. This can arise due to the intrinsic probabilistic nature of $Y|X$, which a deterministic predictor $h$ cannot handle (e.g., coin toss), or the training data containing too many errors (e.g., incorrectly labeled examples), which would make it difficult \changed{for $h$ to approximate $f$}. 
In classification, this can arise due to classes overlapping in certain regions of the instance-space (Figure~\ref{fig:exampleclass2}a), while in regression, \changed{it leads to high variance} in the target variable in certain regions (Figure~\ref{fig:examplereg2}a). One way to interpret this issue is that the chosen feature space does not allow for accurately determining the target value (e.g., missing features might cause examples with different predictions to be projected onto the same example~\citep[Figure~2]{VanCraenendonck2018}).
}

\changed{Second, a poor choice of the predictor's hypothesis space makes it impossible to learn the relationship between $X$ and $Y$.
This occurs when the chosen hypothesis space $\mathcal{H}$ does not include $f$. Figure~\ref{fig:exampleclass2}b illustrates this error in binary classification, where only linear models (e.g., perceptron) are considered for $\mathcal{H}$ while the true target concept $f$ is non-linear. Similarly, Figure~\ref{fig:examplereg2}b shows a regression problem with $\mathcal{H}$ as the family of quadratic models, while $f$ is non-quadratic. In both cases, the expected prediction error is large for certain examples because $f \not\in \mathcal{H}$.
}

\subsubsection{Novelty Rejection}
Many machine learning models struggle when forced to extrapolate to regions of the feature space that were not (sufficiently) present in the training data~\changed{\citep{Cordella1995,SambuSeo2000,Vailaya2000}}.

\changed{\textit{Novelty rejection} allows a model to abstain from making predictions for examples that are sufficiently different from the training data~\citep{Dubuisson1993}. For such examples, the predictor $h$ is likely to make mistakes because the absence of training data similar to $x$ prevents $h$ from learning the correct target $y$.}

\changed{In practice, this arises when one of the following assumptions of the training procedure is violated.} First, the sampling distribution differs from the true distribution. Thus, parts of the feature space are not represented in the data (e.g., the data does not contain any examples of patients suffering from a particular rare disease)~\citep{VanderPlas2021} or one cannot generate data for all imaginable situations (e.g., a sensor can break in many different ways)~\citep{Hendrickx2022,Urahama1995,Hsu2020}. Second, the skew in the class distribution is too large that the model ignores parts of the feature space. Thus, the predictor may choose to ignore these examples in its objective to optimize the accuracy model-complexity trade-off. Third, a new distribution appears after training and these new examples are out-of-distribution. For instance, this can arise due to drift that leads to new classes~\citep{Landgrebe2004} or an adversarial agent that deliberately tries to mislead~\citep{Wang2020,Corbiere2022}.

\begin{figure}
\includegraphics[width=\linewidth]{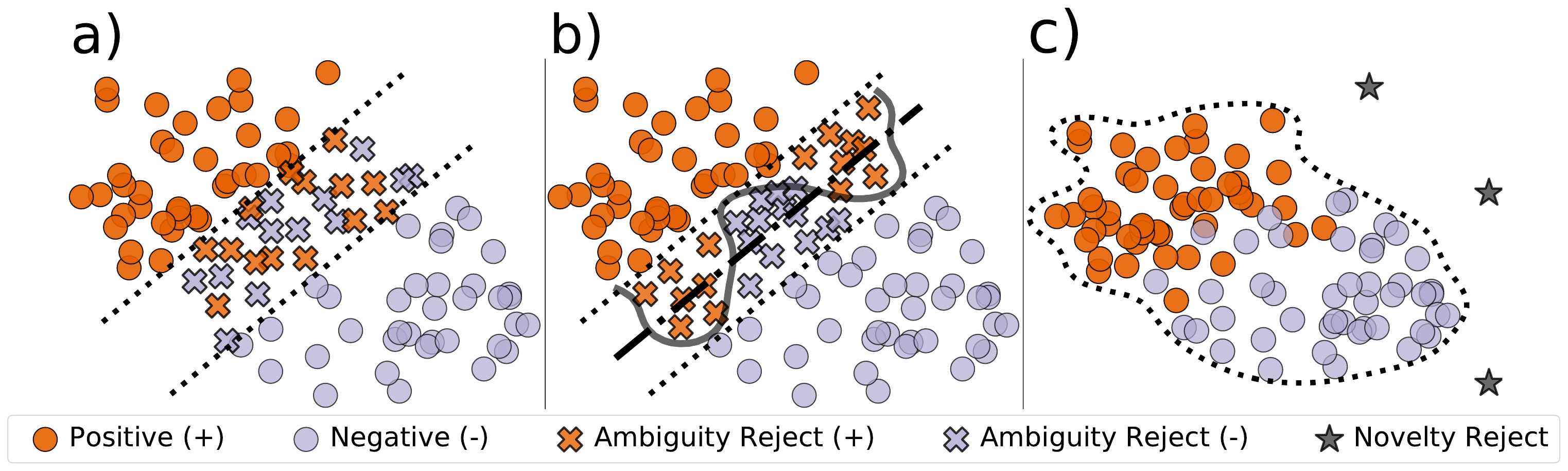}
\caption{Illustration of a classification scenario where the two classes overlap in a region. The dotted lines represent the rejector, the dash-dotted line the fitted predictor $h$, and the solid line the ground truth relation $f$. a) shows ambiguity rejection due to a non-deterministic relation between $X$ and $Y$, b) introduces ambiguity rejection due to the model bias, and c) illustrates an example of novelty rejection. While in the first two plots the rejection region is inside the two dotted lines (examples with cross marks are rejected), in the third figure the rejected novel examples (stars) are outside the dotted line.
}
\label{fig:exampleclass2}
\end{figure}

\begin{figure}
\includegraphics[width=\linewidth]{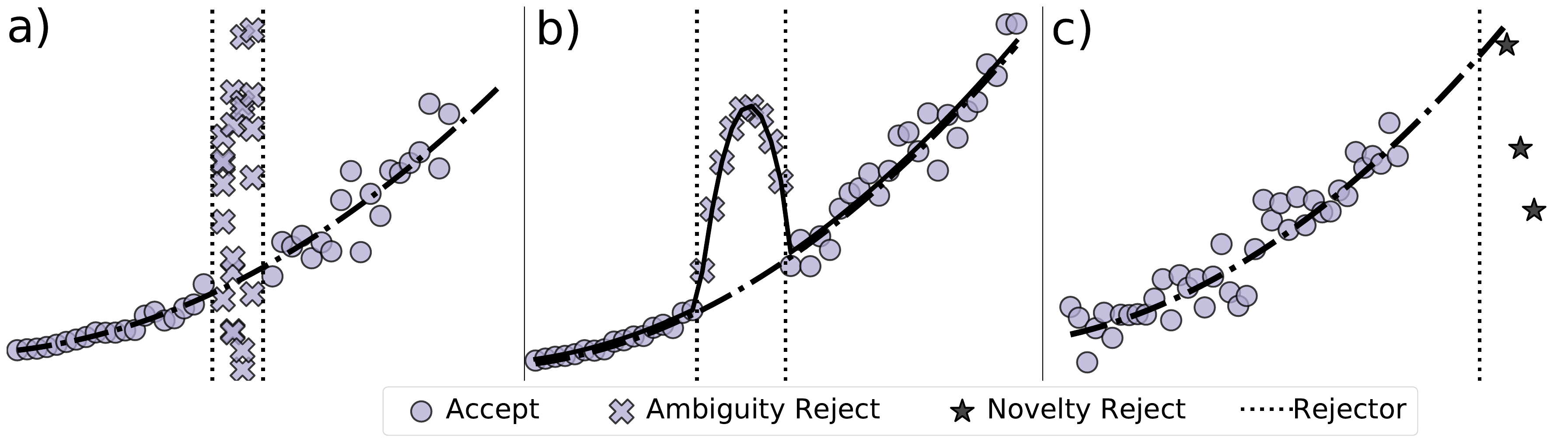}
\caption{Illustration of a regression scenario where rejection can be applied. The dash-dotted line represents the predictor $h$, while the solid line indicates the true function $f$. In a), examples in between the dotted lines are rejected (cross mark) due to the high variance of $Y$. In b), examples indicated with the cross mark are rejected due to the incorrect model bias. In c), star-marked examples are rejected because they differ from the training data (novelties).
}
\label{fig:examplereg2}
\end{figure}

Figures~\ref{fig:exampleclass2}c and~\ref{fig:examplereg2}c illustrate this rejection type for a classification and a regression problem. In both cases, the three black stars represent test examples that are far away from the training examples. 
The learned model $h$ may be inaccurate in these regions due to the lack of training data, which increases the chance of making a misprediction for these test examples.

\begin{figure}
    \includegraphics[width=\textwidth]{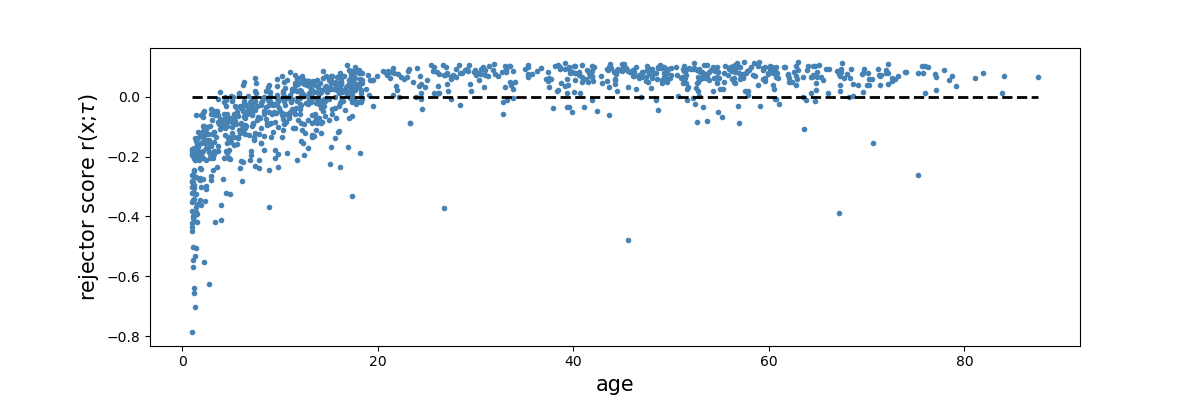}
    \caption{Results of a novelty reject option in a sleep stage scoring application \citep{VanderPlas2021}. The predictor is trained only on adults. The accuracy on children ($66.8\%$) is much lower than the accuracy on adults ($77.7\%$). Introducing the reject option mitigates the risk of making incorrect predictions as it ends up rejecting the predictions for most children.}
    \label{fig:sleep_example}
\end{figure}

The importance of \changed{novelty rejection} often becomes particularly apparent in medical applications since in many of these applications, it is challenging or expensive to get an exhaustive training set. Therefore, the training set might only include patients from specific age groups or with particular medical conditions. 
For instance, \citet{VanderPlas2021} learned a sleep stage classifier but only had access to a training set containing adult patients. Consequently, the model may not perform well when applied in practice to patients that are different than those encountered during training such as children or people suffering from extremely rare disorders. However, this information and the assumptions made during data collection might not be available to the user of the model. 
Therefore, adding a novelty rejector is crucial to avoid poor prediction performance on these patients. In this case, a rejector based on a LOF outlier detector can reject the predictions from children as they have a different morphology than the adults in the training set (Figure~\ref{fig:sleep_example}). This mitigates the risk of incorrect predictions.

%% file: c_metrics.tex
\newpage
\section{Evaluating models with rejection}~\label{sec:metrics}

\begin{table}[htbp]
\centering
\normalsize{
\begin{tabular}{cc|c|c}
\multicolumn{2}{c}{}  & \multicolumn{2}{c}{\textbf{Rejection}} \\
\parbox[t]{1mm}{\multirow{2}{*}{\rotatebox[origin=c]{90}{\textbf{Prediction}}}}                            &           & \emph{No}                & \emph{Yes}               \\
                            \cmidrule(lr){3-4}
                  & \emph{Correct}   & True accept (TA)  & False reject (FR) \\
                            & \emph{Incorrect} & False accept (FA) & True reject (TR) 
\end{tabular}
}
\vspace{0.3cm}
\caption{Confusion matrix for learning-to-reject models. The columns show whether an example is rejected, the rows show whether $h$'s prediction is correct.}\label{tab:confusion_matrix_for_rejection}
\end{table}

\noindent Conceptually, a model with a reject option involves evaluating the outputs of both the predictor and the rejector. Thus, its performance can be viewed through the prism of the confusion matrix shown in Table~\ref{tab:confusion_matrix_for_rejection}, where the columns represent the rejector's decision and the rows whether the predictor's output is correct or not.
Intuitively, the learning-to-reject model is ``correct'' if a correct prediction is returned to the user (a true accept) or an example is rejected when the model's prediction was wrong (a true reject). It is considered to have made a ``mistake'' if the model provides the user with an incorrect prediction (false accept) or rejects an example for which the model's prediction was correct.

Viewed through this lens, a model with rejection has two goals. On the one hand, it wants to have high accuracy $\mathcal{A}$ on examples for which it makes a prediction:
\begin{eqnarray*}
\mathcal{A} &=& \frac{TA}{TA+FA}.
\end{eqnarray*}
\noindent This makes the model reliable as practitioners can trust its outputs. On the other hand, it wants to have high coverage:
\begin{eqnarray*}
    \phi &=& \frac{TA+FA}{TA+FA+FR+TR}.
\end{eqnarray*}
\noindent That is, it should make a prediction for as many test examples as possible~\citep{DeStefano2000,El-Yaniv2010,Lei2014}. This can alternatively be viewed as having a low rejection rate, defined as $1-\phi$. 
This makes the model useful in practice as its predictions can be utilized for decision-making. Unfortunately, these two goals are competing as the accuracy can be increased by limiting the predictions to the most confident cases, i.e., reducing the coverage~\citep{Hansen1997a,Homenda2016}. As a result, metrics specifically tailored to learning to reject must capture this trade-off~\citep{Cordella}. 

Broadly speaking, three categories of metrics exist that evaluate different aspects of a model with rejection:
\begin{description}
    \item[\emph{Evaluating models for a given rejection rate}:] This entails having a fixed rejection rate provided by the user. In this case, one only needs to evaluate performance on the non-rejected examples and it is possible to use the standard evaluations (e.g., accuracy~\citep{Golfarelli1997}); 
    \item[\emph{Evaluating the overall model performance/rejection trade-off}:] One can plot the rejection rate on the x-axis and the predictor's performance obtained on a representative test set on the y-axis, similar to \gls{roc} analysis;
    \item[\emph{Evaluating models through a cost function}:]
    This case only requires knowing the model's output and the costs for (mis)predictions and rejections.
\end{description}

\subsection{Evaluating models with a fixed rejection rate.}
Given a dataset and a fixed rejection rate, \citet{Condessa2017} argue that a good evaluation metric should meet four main criteria: given a fixed predictor, such metric should: \textbf{(p1)} depend on the model's rejection rate; \textbf{(p2)} be able to compare two models with different rejectors for a given rejection rate (and for the same predictor); \textbf{(p3)} be able to compare two models with different rejectors with different rejection rates when one clearly outperforms the other; \textbf{(p4)} reach its maximum value for a perfect rejector (i.e., a rejector that rejects all misclassified examples) and its minimum value for a rejector that rejects all accurate predictions. In addition, \citet{Condessa2017} propose three types of evaluation metrics that meet the required conditions:

\begin{enumerate}
\item[\textbf{a)}] \textbf{Prediction quality.} The model's \textit{prediction quality} (PQ) measures the predictor's performance on the non-rejected examples. For instance, one can use classical evaluation metrics on the accepted examples such as the accuracy
\begin{eqnarray*}
   PQ_{\textsc{acc}}&=& \frac{TA}{TA+FA},
\end{eqnarray*}
the F-scores~\citep{Pillai2011,Mesquita2016} or any other evaluation metric, including fairness metrics~\citep{Madras2018}.
While this allows comparing models with different rejectors, looking only at the prediction quality will tend to  favor the model with the highest rejection rate. By being more conservative (i.e., having a lower coverage), the model tends only to offer predictions for the subset of examples for which it is most confident. 
\item[\textbf{b)}] \textbf{Rejection quality.} The \textit{rejection quality} (RQ) indicates the rejector's ability to reject misclassified examples. 
One way to do this is by comparing the ratio of misclassified examples on the rejected subset ($\frac{TR}{FR}$) to the ratio on the complete dataset ($\frac{FA+TR}{TA+FR}$), i.e. 
\begin{eqnarray*}
   RQ_{\textsc{ratio}} &=& \frac{TR}{FR} \Big/ \frac{FA+TR}{TA+FR}.
\end{eqnarray*}
Looking only at the rejection quality will favor models with the lowest rejection rate. The lower the rejection rate is, the more likely the rejector is to abstain only on those few examples for which it is most confident that the predictor will make a mistake. 
\item[\textbf{c)}] \textbf{Combined quality.} The \textit{combined quality} (CQ) evaluates the model with a reject option as a whole. 
One way to accomplish this is by combining the predictor's performance on the non-rejected examples (prediction quality) with the rejector's performance on the misclassified examples (rejection quality). For instance, using $PQ_{\textsc{acc}}$ and $RQ_{\textsc{ratio}}$ yields to 
\begin{eqnarray*}
   CQ_{\textsc{acc-ratio}} &=& \frac{TA + TR}{TA+FA+FR+TR}.
\end{eqnarray*}
Overall, the combined quality offers a more holistic assessment of the model's overall performance as it measures both the predictor's and the rejector's quality~\citep{Lin2018}.
The downside is that aggregating the two metrics yields a less fine-grained characterization of the model performance. Specifically, in case a model has low CQ, it is hard to ascertain which component, the predictor or rejector, is contributing the most to the model's poor performance.
\end{enumerate}

\paragraph{Pros and Cons.} The main advantage of this category is that the metrics clearly measure the fine-grained model performance in the given setting. However, using one of these types of metrics may be limiting in some cases. For instance, theoretical research may not care about evaluating the model with rejection for a specific rejection rate, as it is usually specified based on domain knowledge. Moreover, given a rejection rate, not all performance metrics can be always used, as some may suffer from task-related issues. For instance, rejecting a whole class would not allow utilizing metrics like F1-score and AUC for the prediction quality as they need both classes' labels.

\subsection{Evaluating the model performance/rejection trade-off.}
\label{subsec:eval_arc}

\begin{figure}[htbp]
    \centering
    \includegraphics[width = \textwidth, trim={30 0 30 35}, clip]{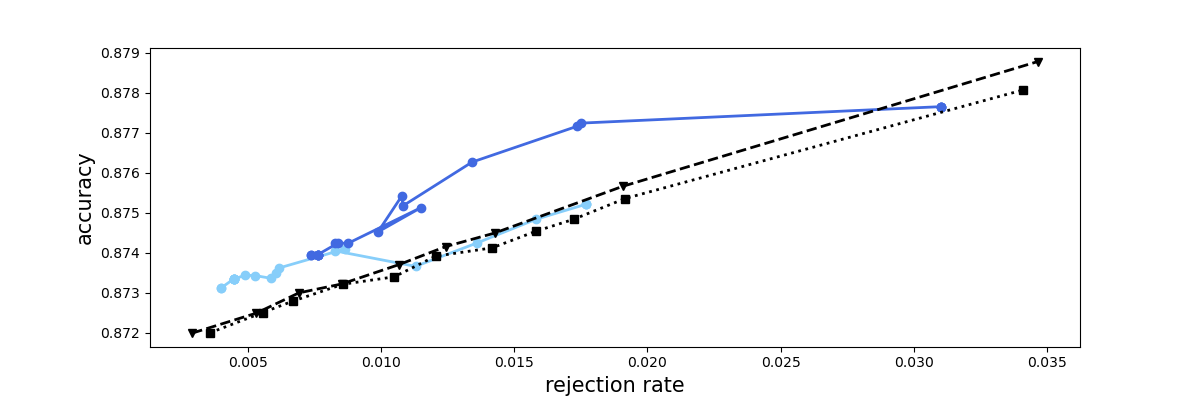}
    \caption{Example of an accuracy-reject curve in which \citet{VanderPlas2023} compare two of their proposed models with a \changed{novelty} reject option (blue) with two models that were used as baseline (black). The proposed models outperform the baselines but the light blue model only does for rejection rates lower than $0.01$.}
    \label{fig:arc_metric}
\end{figure}

To assess the performance-rejection trade-off, a common approach is to evaluate the prediction quality (for non-rejected examples) by varying the rejection rate from $0\%$ to $100\%$, which is known as the \gls{arc}~\citep{Nadeem2010a}. This involves plotting the rejection rate on the x-axis and the prediction quality (e.g., accuracy) on the y-axis~\citep{Hanczar2014}. Higher curves indicate better performance. Alternatively, risk-based metrics like mean squared error can be plotted, with lower curves indicating better performance~\citep{SambuSeo2000}. Sometimes the predictor's performance on the rejected examples is also shown with the intuition that it should be worse on this subset of the data than on the accepted ones~\citep{Zou2011,Condessa2015,Condessa2015b}.

When comparing two models using the \gls{arc} method, two scenarios arise. First, one model clearly outperforms the other in terms of prediction quality for all rejection rates. Second, the models show varying prediction quality across different rejection rates. To illustrate this, consider Figure~\ref{fig:arc_metric} where the light blue curve  only outperforms the black curves if the rejection rate is lower than $0.01$. When it is not clear which model performs best, the overall performance can be assessed using the \gls{aurc}, similar to the AUC in standard machine learning~\citep{Vanderlooy2006a,Vanderlooy2006,Landgrebe2006}. In this example, the \gls{aurc} of the light blue curve will be higher than the \gls{aurc} of the black curves, indicating that it performs better overall than the black curves.

\paragraph{Pros and Cons.} The main advantage of this category is that any prediction quality metric can be used on the y-axis. Moreover, they provide a high-level overview of how the model works for different rejection rates. However, generating these curves can be challenging for two reasons. 
First, some rejectors do not allow directly setting the rejection rate~\citep{Wu2007,Homenda2014}, but they might have other hyperparameters (e.g., rejection threshold). 
Mapping these to rejection rates may be challenging and may not be possible to achieve all possible rejection rates.
Second, altering the rejection rate of a model with rejection can require completely retraining the whole model. This may be too computationally demanding to perform a fine-grained analysis~\citep{Condessa2017}.

\subsection{Evaluating models through a cost function.}
\label{sec:eval_arc}
For classification tasks, one can ask the user to specify the costs for (mis)predictions as well as for rejection and evaluate a model by its (expected) total cost at test time.
Although the costs can be designed as a continuous function of the examples~\citep{Mozannar2020,De2021}, the cost function typically accounts only for three constant costs: the cost of correct prediction $C_c$, the cost of a prediction error $C_e$, and the cost of rejection $C_r$ such that $C_c < C_r < C_e$~\citep{DeStefano2000,Balsubramani2016,Condessa2013}. \changed{Without loss of generality, usually one assumes normalized costs, i.e., $C_c = 0$, $C_e = 1$ and $C_r\in [0,1]$ in which the normalized value for $C_r$ can be obtained from the initial values as $\frac{C_r - C_c}{C_e - C_c}$.} Although the costs need to be set based on some domain knowledge, there are two constraints for setting the rejection cost properly. First, it must be lower than a random predictor's average cost for a classification task with $K$ classes, i.e. $C_r \le \frac{1}{K}$~\citep{Cordella1995a,Herbei2006c}. Otherwise, the expected cost of always making a random prediction would be lower than the rejection cost, which nullifies the task of rejection. Second, one should account for possible imbalance classes by setting $C_r \le 1 - \max_{k \le K} P(Y = k)$, where $P(Y = k)$ is the class frequency~\citep{Perini2023}. In fact, for higher $C_r$, a naive model that always predicts the most frequent class $\bar{k}$ would obtain a cost equal to $1 - P(Y = \bar{k})$ and rejecting examples would not be worth it for higher rejection costs.

\paragraph{Pros and Cons.} The main benefit of this category is its high interpretability: given a final cost, we can easily go back to the causes that yield such a cost. Moreover, one can use the same cost function to optimize the model parameters during the learning phase. This ensures coherence between learning the optimal model at training time and measuring its performance at test time. However, this category has the key drawback that the user must set the cost function based on domain knowledge, which is not always easy to obtain. Setting different costs changes the quality of the models, which may end up ranking several compared models differently.
A way to alleviate this would be to make the Cost-Reject plot~\citep{Hanczar2019}, where, similar to the ARC, the x-axis and y-axis represent respectively the normalized rejection cost and the normalized prediction cost~\citep{Friedel2006,Abbas2019}.

%% file: d_SeparatedRejector.tex
\section{Separated rejector}\label{sec:separated_rejector}

Separated rejectors operate by filtering out unlikely examples. They are typically used for novelty rejection though there are some examples of using them for ambiguity rejection~\citep{Asif2020}. 
Because the rejector does not use the predictor's output in any way, it is typically a function of the examples: $r \colon \mathcal{X} \to \mathbb{R}$. Formally, the separated architecture yields the following model:
\begin{equation}
m(x) = 
\begin{cases}
\rr{} & \text{if } r(x) < \tau \\
h(x)	  & \text{otherwise.}
\end{cases}
\label{eq:separated_rejector}
\end{equation}
If the rejector $r$ outputs a value less than $\tau$, then the model $m$ rejects the example ($\rr{}$). Otherwise, $m$ uses the predictor $h$ to make a prediction.  

The separation between predictor and rejector means that the rejector is learned independently of the predictor. The learning task involves learning the rejector itself as well as setting the threshold $\tau$. To align with its goal of identifying unlikely or unexpected examples, a common choice is to use anomaly/outlier, out-of-distribution, or novelty detection algorithms. Three categories of methods can be used for this goal: models that 1) estimate $p(X)$, 2) are one-class classifiers, and 3) quantify the degree of novelty using a data-driven score function.

\paragraph{\textbf{Learning a separated rejector.}} 
A first option to learn a separated rejector is to use a probabilistic model that estimates the marginal density $p(X)$ and to reject a test example $x$ if $p(x)<\tau$. These probabilistic models often make assumptions about the distribution of the examples and are trained to maximize the likelihood of the training dataset~\citep{Vasconcelos1993}. For instance, \citet{Landgrebe2004} proposes a locally normal distribution assumption and uses a Gaussian Mixture Model (GMM) to estimate $p(X)$ with a specified number of components whereas others have considered Variational Autoencoders~\citep{Wang2020} and Normalizing Flows~\citep{Nalisnick2019}.

Another option to learn a separated rejector is by employing a one-class classification model. Generally, they enclose the dataset into a specific surface and flag any example that falls outside such region as novelty. For instance, a typical approach is to use a \gls{ocsvm} to encapsulate the training data through a hypersphere~\citep{Coenen2020,Homenda2014}. By adjusting the size of the hypersphere, the proportion of non-rejected examples can be increased~\citep{Wu2007}. 

Alternatively, some models assign scores that represent the degree of novelty of each example (i.e., the higher the more novel), such as $LOF$~\citep{VanderPlas2023} or Neural Networks~\citep{Hsu2020}. When dealing with these methods, one often initially transforms the scores into novelty probabilities using heuristic functions, such as sigmoid and squashing~\citep{Vercruyssen2018}, or Gaussian Processes~\citep{Martens}. Then, the rejection threshold can be set to reject examples with high novelty probability.

\paragraph{\textbf{Learning the rejection threshold $\tau$.}} The rejection threshold $\tau$ is a crucial parameter that determines whether an example is rejected or not. In many cases, the threshold is set based on domain knowledge. For instance, one can introduce adversarial examples and set a threshold to reject them all~\citep{Hosseini2017}. In case the number of novelties is unknown, one can use existing methods to estimate the contamination factor, i.e. the proportion of novelties, and set the threshold accordingly~\citep{Perini2022,Perini2022a}. Otherwise, heuristics can be employed, such as rejecting examples falling within the first or second percentiles of correctly classified training examples~\citep{Wang2020}. 

\paragraph{\bf{Benefits and drawbacks of a separated rejector.}} Using a separated rejector has several \emph{benefits}. First, the rejector is predictor agnostic. Hence, it can be combined with any type of predictor. Second, because the rejector can be trained independently of the predictor, it is possible to augment an existing predictor with a reject option using this architecture. 
Third, by serving as a filter, the predictor makes fewer predictions. This is particularly advantageous when there is a high computational cost associated with using the predictor.
Finally, this architecture is generally simpler to operationalize compared to rejectors that interact with the predictor.
However, there are two evident \emph{drawbacks}. First, not sharing information between the predictor and the rejector results often in sub-optimal rejection performance~\citep{Homenda2014}. 
Second, this architecture is typically only used for novelty rejection because it is naturally related to assessing whether $x$ is rare or not while ambiguity rejection requires information on $p(Y|X)$, which is often estimated through the predictor's output.

%% file: e_DependentRejector.tex
\section{Dependent rejector}\label{sec:dependent_rejector}
Dependent rejectors analyze the predictor's output to identify examples that the predictor is likely to mispredict. The rejector is typically represented as a confidence function $c_h \colon \mathcal{X} \to [0,1]$ that measures how likely the predictor is to make a correct prediction. Formally, the model for a dependent architecture has the form:
\begin{equation}
m(x) = 
\begin{cases}
\rr{} & \text{if } r(x; h) < \tau; \\
h(x)  & \text{otherwise.}
\end{cases}
\label{eq:dependent_rejector}
\end{equation}
where $r$ may depend on the feature vector $x$, and on the predictor $h$ through the confidence function $c_h$. Similar to the separated rejector, the model rejects the example ($\rr{}$) only if the rejector outputs a value lower than $\tau$. Without loss of generality, we assume the confidence values $c_h(x) \in [0,1]$, as one can always transform a score function into this range.  

The learning task for a dependent rejector usually entails (1) selecting a confidence function that measures how confident the predictor is in its predictions, and (2) setting the rejection threshold $\tau$. 

\paragraph{\textbf{Learning a dependent rejector: types of confidence function $c_h$.}}
The form of  the confidence function depends on the desired rejection type. For \emph{ambiguity rejection}, the metric should indicate the variability of the target variable or the potential predictor bias. For \emph{novelty rejection}, the confidence should capture the example's similarity to the training data. 
We distinguish among four ways to derive the confidence scores: i) estimating the conditional probability $P(Y|X)$, ii) estimating the class conditional density $p(X|Y)$, iii) performing a sensitivity analysis, and iv) exploiting the predictor's properties.

The \emph{conditional probability} approaches allow for ambiguity rejections by using the maximum of the class conditional probability as confidence function~\citep{Pazzani1994,Fumera2004a,Lam1995}
\begin{eqnarray}\label{eq:c_h_class_conditional_ambiguity}
    c_h(x) &=& \max_{k\in \mathcal{Y}} P(Y=k|X=x)
\end{eqnarray}
where $k$ is either the true target value or the predictor's output. Low $P(Y|X)$ values for two or more targets $k_1, k_2 \in \mathcal{Y}$ indicate high randomness of the data or proximity to the decision boundary~\citep{Arlandis2002}. Deriving the conditional probabilities from the predictor's outputs can be done by post-processing the predictor's output using techniques such as sigmoid calibration for binary classification tasks~\citep{Cordella2014,Brinkrolf2017}
\begin{eqnarray*}
    P(Y = 1|X=x) &\approx& \frac{1}{1+\exp(A h(x) + B)},
\end{eqnarray*}
with parameters $A$ and $B$ learned during the training~\citep{Brinkrolf2018}, softmax transformation for multi-class classification tasks~\citep{Kwok1999}
\begin{eqnarray*}
    P(Y = k|X=x) &\approx& \frac{\exp{(h_k(x))}}{\sum_{j \in \mathcal{Y}} \exp{(h_j(x))}},
\end{eqnarray*}
where $h_j(x)$ is the predictor's output for $x$ related to class $j$, or by fitting Gaussian Processes for regression tasks~\citep{SambuSeo2000}. In case of multiple predictors $h_1, \dots, h_V$, one can also measure the ensemble agreement as the conditional probability~\citep{Glodek2012,Zhang2013}
\begin{eqnarray*}
    P(Y = k|X=x) &\approx& \frac{\sum_{v=1}^V \mathbbm{1} [h_v(x) = k]}{V}.
\end{eqnarray*}

On the other hand, the \emph{class conditional density} approaches perform novelty rejection putting~\citep{Dubuisson1993,Dubuisson1985}
\begin{eqnarray}\label{eq:c_h_class_conditional_novelty}
    c_h(x) &=& \max_{k \in \mathcal{Y}} p(X=x|Y=k).
\end{eqnarray}
Intuitively, a low density $p(X|Y)$ expresses that a sample is rare~\citep{Condessa2015b}. Common methods to estimate the confidence in Eq.~\ref{eq:c_h_class_conditional_novelty} employ generative predictors that directly measure the data density such as \glspl{gmm}~\citep{Vailaya2000}. It is also possible to employ heuristic approaches such as normalizing the class distance between the example $x$ and its $v-$th nearest neighbor $x'$~\citep{Conte2012,Villmann2015,Fischer2014a}, i.e.
\begin{eqnarray*}
    p(X=x|Y=k) &\approx& \frac{d(x,x')}{\sum_{\{(x_*, y_*) \in D \colon y_* = k\}} d(x_*,x'_*)}
\end{eqnarray*}
and computing the proportion of neighbors within a specified radius $R$~\citep{Berlemont2015} as 
\begin{eqnarray*}
    p(X=x|Y=k) &\approx& \frac{|\{x' \colon d(x, x') \le R, (x',y')\in D, y' = k\}|}{\sum_{(x',y') \in D} |\{x_* \colon d(x', x_*) \le R, (x_*,y_*)\in D, y_* = k\}|}.
\end{eqnarray*}

The \emph{sensitivity analysis line} allows only \emph{ambiguity rejection}, as it measures the robustness of the predictor under perturbation of either (a) its parameters or (b) the examples~\citep{Lewicke2008,Hellman1970}. 
Intuitively, slightly perturbing the predictor's parameters has major effects on the predictions only for examples that fall in the proximity of the decision boundary: slight variations of the parameters yield slight changes in the decision boundary, which, in turn, may end up flipping the predictions for some examples. Examples of employed perturbations involve adding some noise to the model's parameter values (e.g., adding a random sample from a normal distribution with null mean and small variance to the weights of a neural network), employing neural networks with a dropout layer~\citep{NIPS2017_7073}, or using a Bayesian simulation~\citep{Perini2020}.
In the case of constructing multiple predictors, a common and simple confidence metric is
\begin{equation*}
    c_{\{h_1,\dots, h_M\}} (x) = 1 - \Var\{h_1(x),\dots,h_M(x)\}
\end{equation*}
where $h_1, \dots, h_M$ are the $M$ predictors constructed by perturbing the parameters, and the variance is scaled to be in $[0,1]$. In some cases, one can directly employ an ensemble of $M$ similar predictors and measure the variance of their predictions~\citep{Fumera2004,Jiang2020}.

Alternatively, one can perturb the test example $x$ to be $x+\varepsilon$, where $\varepsilon$ is a random noise such that $\|\varepsilon\|$ is small. Intuitively, we want a predictor's output to remain the same when the example is only slightly perturbed. Thus, the confidence metric should reflect the robustness of $h$~\citep{Mena2020,Denis2020,Joana2021}, such as
\begin{equation*}
    c_{h} (x) = P(h(x+\varepsilon) = h(x)),
\end{equation*}
which measures how likely it is that the prediction does not change when the example is perturbed. More generally, we can apply transformations such as rotations and symmetries to the examples~\citep{Chen2018}. Finally, because choosing a specific value for $\varepsilon$ is hard, one can use existing approaches to find each example's minimum $\varepsilon$ that will alter its predicted label~\citep{Devos2021} and derive a confidence metric as a function of the training $\varepsilon$.

The \emph{property-based} methods consist of learning the confidence based on some of the predictor's properties, such as using the leaf configurations of a tree ensemble or the neural network's weight of specific neurons. This line allows both ambiguity and novelty rejection, depending on the utilized property. These methods tend to exploit heuristic and data-driven intuitions and there are no overarching themes that connect these intuitions. 
For instance, \citet{Devos2023} present a method to detect evasion attacks in tree ensembles. By enumerating the leaves of each tree as $o_i$, they map each example $x$ to the configuration $o = (o_1,\dots,o_V) \in \mathbb{N}^V$ of the $V$ activated leaves (one per tree) when passing $x$ as input to the ensemble. In such output configuration space, they quantify the proximity to the decision boundary by measuring the Hamming distance between the configuration $o$ of a test example with ensemble's prediction $y$ and the closest training example's configuration $o'$ with flipped prediction $\hat{y} \ne y$:
\begin{equation*}
    \textsc{OC-score}(x) = \min_{o' \in R_{\hat{y}}} \left(\sum_{v=1}^V \mathbbm{1} \left[o_v \ne o'_v\right]\right)
\end{equation*}
where $R_{\hat{y}}$ is the set of training configurations with flipped predictions. One can derive a confidence metric $c_h$ by, for instance, min-max normalizing the $\textsc{OC-score}$.
On the other hand, confidence values can also be derived from the weight vectors of a \gls{som}~\citep{Sousa2014,GamelasSousa2015}. Because \gls{som}’s can approximate the input data density, they approximate $p(X=x|Y=k)$ with $p(w|Y=k,X=x)$, where $w$ are the weights of the neural network, using standard statistical techniques, such as the Parzen Windows~\citep{Vesanto1999}.
Moreover, \citet{El-Yaniv2011} propose a disbelief principle, which computes the confidence function by measuring how much the predictor $h$ deteriorates if retrained with the constraint to predict a specific example $x$ differently (i.e., $h_x$)
\begin{eqnarray*}
    c_h(x) = \frac{1}{R(h_x) - R(h)},
\end{eqnarray*}
where $R(h_x) > R(h) > 0$.
Finally, the literature presents additional ad-hoc confidence metrics for $k$-NN and Random Forest~\citep{Gopfert2018,Dalitz2009}.

\paragraph{\bf{Learning the rejection threshold $\tau$.}}

Setting an appropriate rejection threshold $\tau$ is crucial for having an accurate dependent rejector. At a high level, the threshold is set in three main ways: using domain knowledge,  adhering to user-provided constraints, or tuning it empirically based on some objective function.

In some situations, users possess \emph{domain knowledge} that enables setting $\tau$ to achieve a desired rejection rate $\rho$~\citep{LeCapitaine2012,Pang2021}. Setting $\tau$ in this situation entails (1) ranking training examples based on their confidence level, and (2) setting $\tau$ such that the desired percentage of predictions are rejected~\citep{Sotgiu2020}, that is, at
\begin{equation*}
    P_X(c_h(x) < \tau) = \rho.
\end{equation*}
Note that this approach is also used when evaluating the model performance/rejection trade-off, which needs to measure the model performance for a fixed rejection rate (see Sec.~\ref{subsec:eval_arc})~\citep{Ma2001,NEURIPS2018_285e19f2,Fumera2003}.

In other cases, users provide knowledge as \emph{specific constraints} that should be satisfied~\citep{Pietraszek2005b}. On the one hand, the user may provide an upper bound $\mathfrak{R}$ for the rejection rate and aim to limit the number of rejections. This results in learning the appropriate $\tau$ by minimizing the model misclassification risk while adhering to the rejection rate constraint~\citep{Zhou2022,Pugnana2023a,Pugnana2023}
\begin{equation*}
    \tau = \argmin_{t \in [0,1]} P_{XY}(h(x) \ne y, c_h(x) \ge t)  \quad \text{ subject to } \quad P_X(c_h(x) < t) \le \mathfrak{R}.
\end{equation*}
On the other hand, the user may provide an upper bound $\mathfrak{M}$ for the proportion of mispredictions, and aim to control the allowable error~\citep{Varshney2011,Sayedi2010}. Thus, one needs to learn $\tau$ by setting up the complementary problem as before, namely by minimizing the model rejection rate while satisfying the constraints on error~\citep{Li2006,Franc2019,Franc2021}
\begin{equation*}
    \tau = \argmin_{t \in [0,1]} P_X(c_h(x) < t) \quad \text{ subject to } \quad P_{XY}(h(x) \ne y, c_h(x) \ge t) \le \mathfrak{M}.
\end{equation*}
Moreover, one can generalize this problem by finding the threshold $\tau$ such that the predictor’s misclassification risk at test time is guaranteed to be bounded with high probability~\citep{NIPS2017_7073}.

Finally, it is possible to set $\tau$ empirically according to some objective function. 
The most common approach is to set a \emph{single global threshold} $\tau$, which makes the rejection both simple and transparent, yet usually effective~\citep{Fukunaga1972}. 
This is the case for Chow's rule \citep{Chow1970a} which involves learning the optimal $\tau$ by minimizing the risk function that includes the expected error rate and the rejection rate
\begin{equation}\label{eq:chows_minimumrisk_rule}
    \tau = \argmin_{t\in[0,1]} \left[{\underbrace{\int_{\{x \in \mathcal{X} \colon c_h(x) \ge t\}} (1 - c_h(x)) \, p(x) \, dx}_{\textsc{Error rate}}} + t {\underbrace{\int_{\{x \in \mathcal{X} \colon c_h(x) < t\}} p(x) \, dx}_{\textsc{Rejection rate}}}\right].
\end{equation}
However, in real-world scenarios, obtaining complete knowledge of class distributions is challenging, limiting the applicability of Chow's rule~\citep{Shekhar2019}. 
Thus, in a binary classification case, \citet{Tortorella2000} proposes to use two rejection thresholds $\tau_1$ and $\tau_2$ such that
\begin{equation}
h(x) = 
\begin{cases}
0 & \text{if } c_h(x) < \tau_1; \\
1  & \text{if } c_h(x) > \tau_2; \\
\rr{} & \text{if } \tau_1 \le c_h(x) \le \tau_2
\end{cases}
\end{equation}
with $c_h(x) = P(Y=1|X=x)$.
He proposes to learn $\tau_1$, $\tau_2$ by optimizing a cost function that is identical to finding the intersection between  the cost function and the convex hull of the \gls{roc} curve
\begin{eqnarray}
    \tau_1 &=& \argmin_{t \in [0,1]} P(Y = 1) (C_{fn} - C_r) FNR(t) + P(Y=0) (C_{tn}-C_r) TNR(t)\\
    \tau_2 &=& \argmin_{t \in [0,1]} P(Y=1) (C_{tp} - C_r) TPR(t) + P(Y=0) (C_{fp} - C_r) FPR(t)
\end{eqnarray}
where $C_{fn}$, $C_{fp}$, $C_{tn}$, and $C_{tp}$ are the costs for false negatives, false positives, true negatives, and true positives, while $FNR(t)$, $TNR(t)$, $TPR(t)$ and $FPR(t)$ are the false negative, false positive, true negative and true positive rates obtained by evaluating the models with the thresholds set to $t$.
This approach is theoretically equivalent to Chow's rule under the Bayesian optimality assumption~\citep{Santos-Pereira2005,Du2010}. However, when estimating posterior probabilities, Chow's rule is not suitable, and $\tau$ should be learned using a cost-based approach~\citep{Marrocco2007,Kotropoulos2009}.
Different approaches have extended \citeauthor{Tortorella2000}'s method to address other scenarios~\citep{SANSONE2001}, such as stable formulations of ROC curves for small datasets~\citep{Jigang2006}, robust and fast-to-retrain rejections for cost-sensitive situations~\citep{Dubos2016,Fischer2015}, and tailored solutions for learning meta-classifiers or handling multiple classes~\citep{Pietraszek2007,Cecotti2013}.

For tasks requiring a more fine-grained rejection capability, considering \emph{multiple local thresholds} $\tau_1, \tau_2, \ldots$ (up to a finite number) may be beneficial~\citep{Muzzolini1998,Kummert2016,Krawczyk2018}. Normally, setting local thresholds requires dividing the feature space into $J$ regions $\mathcal{J}_i$ and setting a (local) threshold in each region. For instance, one can design regions and thresholds by
\begin{itemize}
    \item constructing one region for each  class, i.e. $\mathcal{J}_i = \{x_* \ | \ y_* = i\}$, which means that the number of regions $J$ equals the number of classes $K$~\citep{Fumera2000a}; then, one often finds the local threshold by using for each $\mathcal{J}_i$ the same approach as for global thresholds;
    \item using the Voronoi-cell decomposition, which requires $J$ prototypes $w_i$ to have $\mathcal{J}_i = \{x_* \ | \ d(x_*, w_j) \le d(x_*, w_k) \ \forall k \ne i\}$, for $i \le J$~\citep{Villmann2015,Fischer2015b,Fischer2015,Fischer2016}; then, \citet{Fischer2014b} present a greedy optimization method to adaptively determine local thresholds using a heuristic principle;
    \item setting up an optimization problem that finds the optimal thresholds by assigning different class rejection costs; for instance, in binary classification, \citet{Zheng2011} proposes to find 
    \begin{equation*}
        \begin{split}
            \tau_1, \tau_2 = \argmin_{0\le t_1, t_2 \le 1} \Big[& P_{XY}(c_h(x)<t_1 | y = 0) C_{r,0} + P_{XY}(c_h(x)<t_2 | y = 1) C_{r,1}\\
            &+ P_{XY}(m(x) \ne y) C_e \Big]
        \end{split}
    \end{equation*}
    where $C_{r,0}$ and $C_{r,1}$ are the costs for rejecting examples from the negative and positive classes;
    \item optimizing an objective function that accounts for different user-specified class misclassification risks $\mathfrak{M}_1, \dots, \mathfrak{M}_K$; \citet{Lin2022} treat each class independently, setting
    \begin{equation*}
        L(t_k) = \hat{A}(t_k) + \sum_j^K \lambda_j (\hat{\mathfrak{M}}_j -  \mathfrak{M}_j)^2
    \end{equation*}
    where $\hat{A}(t_k)$ is any ambiguity metric, $\lambda_j$ is a penalization term that needs to be set to high values to penalize high differences between the model's misclassification risk $\hat{\mathfrak{M}}$ and the user-specified target.
\end{itemize}
Although multiple thresholds give more fine-grained control over a rejector's performance~\citep{Laroui2021,Gangrade2021a}, this is usually more computationally expensive. However, \citet{Fischer2016a} propose efficient schemes for optimizing local thresholds and show that the computation time can be reduced to polynomial~\citep{Boulegane2019}.

\paragraph{\bf{Benefits and drawbacks of a dependent rejector.}} Designing a dependent rejector has several \emph{benefits}.
First, the interaction between the predictor and rejector enables both types of rejection, because the rejector learns from the predictor's output the regions of the feature space where examples are mispredicted or unlikely to fall.
Second, a dependent rejector can extend an existing predictor (including black-box) by simply setting a proper threshold on a confidence measure. Third, it allows the reuse of previously learned models, eliminating the need for costly retraining~\citep{Zou2011,Tang2014}. Fourth, a confidence-based rejection could be improved by considering multiple confidence metrics where each one captures different aspects of the underlying uncertainty~\citep{Tax2008}. However, this architecture has potential \emph{drawbacks} as well. First, the quality of the dependent rejector is highly influenced by the quality of the confidence metric, which is usually hard to evaluate. Second, typically a dependent rejector does not affect the predictor's learning phase. This results in possible sub-optimal predictions of the model with a reject option.

%% file: f_IntegratedRejector.tex
\section{Integrated rejector}~\label{sec:integrated_rejector}

The integrated rejector combines the rejector and predictor into a single model where it is impossible to distinguish between the role of the $h$ and the $r$.  
Formally, the model with integrated reject option acts as
\begin{equation}
    m(x) \in \mathcal{Y} \cup \{\rr{}\}.
	\label{eq:integrated_rejector}
\end{equation}
Conceptually, this model simply includes $\rr{}$ as an additional output.

This architecture usually needs a unique algorithm for learning predictor  and rejector in tandem~\citep{Cortes2016,Cortes2016b}. 
There are two distinct approaches to learning an integrated rejector. The first approach is model-agnostic and involves designing an objective function that penalizes (mis)predictions as well as rejections. The second approach is model-specific and entails integrating a rejector into an existing predictor, where rejection becomes part of the decision-making process.

\paragraph{\bf{Learning a model-agnostic integrated rejector.}}
Typically, learning a model that simultaneously makes accurate predictions and rejects the examples that will be otherwise mispredicted can be done by simply designing a specialized objective function~\citep{Mozannar2020}. Then, such a function can be minimized using potentially any existing optimizer, which makes it model-agnostic.
\changed{For instance, for classification, a simple cost-based objective for any hypothesis $m$ can be expressed as
\begin{equation*}
    L = \E_{p(X,Y)} \left[C_r \mathbbm{1}_{m(x) = \rr{}}(x) + \mathbbm{1}_{m(x) \not\in \{y,\rr{}\}}(x)\right]
\end{equation*}
with $C_r \in (0, 1/2]$.
}
For tasks other than classification, ad-hoc loss functions are used, such as those for multilabel classification~\citep{Pillai2013,Nguyen2020,Nguyen2021}, regression~\citep{Asif2020,Kalai2021}, online learning~\citep{pmlr-v80-cortes18a,Kocak2020}, and multi-instance learning~\citep{Zhang2006}. However, in many cases, surrogate losses are employed to enable efficient optimization, as learning from discrete losses is computationally impractical~\citep{Wegkamp2007,Grandvalet2009,Cao2022}. Consequently, the original loss $L$ is converted into a convex loss by utilizing surrogate functions $\psi \colon \mathbb{R} \to \mathbb{R}$, such as the logistic and hinge functions~\citep{Ramaswamy2018,Zhang2018,Bartlett2008}:
\begin{equation*}
    \text{Logistic: } \psi(L) = \frac{1}{1+ \exp(L)}, \qquad \text{Hinge: } \psi(L) =
    \begin{cases}
        1- \frac{1-C_r}{C_r} L & \text{ if } L<0\\
        1-L  & \text{ if } 0 \le L < 1\\
        0 & \text{ otherwise}
    \end{cases}
\end{equation*}
\changed{For the hinge loss to be convex, it is required that $C_r \leq 1/2$ which is the case for classification as otherwise the cost of rejection would be higher than the cost of random guessing.}
Numerous studies in the literature have explored the properties of surrogate loss functions~\citep{Yuan2010}, including calibration effects~\citep{Ni2019,Charoenphakdee2019}, estimates of bounds for misclassification risk~\citep{Shekhar2019,Kato2020}, penalization effects in high-dimensional spaces~\citep{Wegkamp2012}, proximity to the optimal Bayes solution~\citep{Bounsiar2008,Shen2020}, and convergence rate analysis~\citep{Denis2020}.

Lastly, one can allocate an extra class $K+1$ (commonly known as the \emph{reject class}) for rejection and assign a specific penalization cost $C_r$ for predicting such a class. With this setting, there are two main alternatives. In the first case, there are no actual examples belonging to this class. Thus, these approaches design loss functions to enable the classifier to assign on its own a positive score to ambiguous examples~\citep{Huang2020,Feng2022}. For instance, \citet{Ziyin2019} \changed{propose to measure the expected loss as}
\begin{equation*}
L = \mathbb{E}_{p(X,Y)} \left[\log(s_{k}(x) + \frac{1}{C_r} s_{K+1}(x))\right],
\end{equation*}
where $s_{k}(x)$ and $s_{K+1}(x)$ are probabilities, respectively, for the class $y = k$ and $K+1$ (rejection). At a high level, decreasing the rejection cost $C_r$ results in higher chances of rejection. In the second case, one artificially generates examples ${x_{n+1},\dots,x_{N}}$ (e.g., adversarial examples) and assigns them to the rejection class $K+1$. By training a predictor with $K+1$ classes, the reject option is naturally incorporated as output, and any (multi-class) predictor can be used for novelty~\citep{Vasconcelos1995,Singh2004,Urahama1995} or ambiguity rejection~\citep{Thulasidasan2019,Pang2022}.

\paragraph{\bf{Learning a model-specific integrated rejector.}}

\begin{figure}
\includegraphics[width=\linewidth]{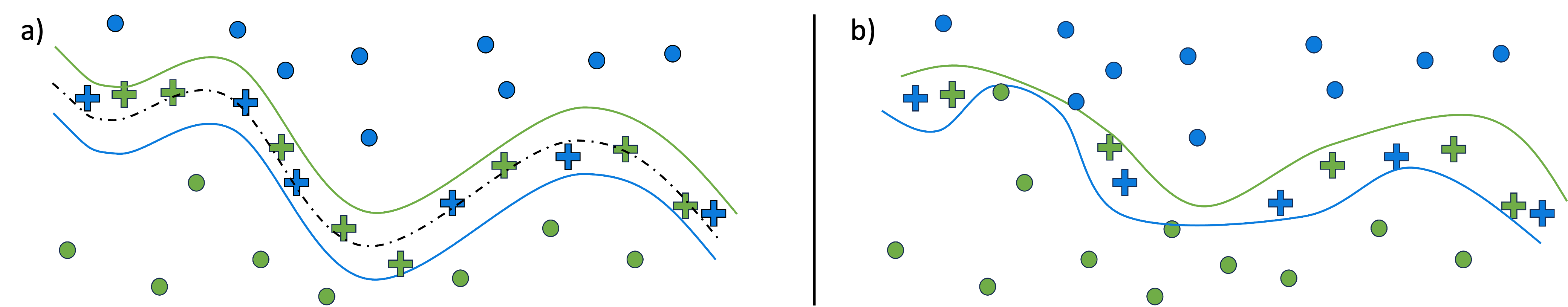}
\caption{Two cases of Integrated SVMs: (a) on the left side, the two hyperplanes are parallel and equidistant to the decision boundary (dashed line); (b) on the right side, each SVM gives higher priority to one class by limiting mispredictions.}
\label{fig:integrated_svm}
\end{figure}

In many practical use cases, one may already know that a specific class of models works well within the given context, such as SVM models in medical applications~\citep{Hanczar2008,Hamid2017}. Given a specific predictor, its learning algorithm can be slightly adapted to include the reject option. 
For instance, \emph{integrated SVMs} set two (or more) hyperplanes on the feature space and reject all the examples located in between them~\citep{Pillai2011,Lin2018,Zidelmal2012HeartbeatCU}. 
Figure~\ref{fig:integrated_svm} shows two common cases to learn the hyperplanes. First, one can parametrize the hyperplane as $w\cdot x + b \pm \varepsilon = 0$, with $\varepsilon\ge 0$, which results in parallel and equidistant hyperplanes from the decision boundary, where $\varepsilon$ indicates the distance~\citep{Fumera2002}. Learning such hyperplanes requires minimizing the empirical loss
\begin{equation*}
    L = \frac{1}{2} w \cdot w + C \sum_{i=1}^n l(\xi_i, \varepsilon) - \sum_{i=1}^n \alpha_i\left[y_i(w\cdot x_i+b)-1+\xi_i\right]
\end{equation*}
where $w$ is the weight vector, $b$ is the intercept of the hyperplane, $C$ is a (large) hyperparameter that regulates the importance of the performance/rejection trade-off expressed inside the function $l(\xi_i,\varepsilon)$, and $\alpha_i$ are the Lagrangian multipliers. Second, the two hyperplanes can be parametrized as 
\begin{equation*}
    w'\cdot x + b' = 0 \quad \text{and} \quad w''\cdot x + b'' = 0.
\end{equation*}
By formulating two distinct optimization problems, one can learn the parameters of these hyperplanes. In this approach, one hyperplane is highly penalized for mispredicting the positive class, while the other one is for the negative class. Essentially, this technique yields two SVMs that have few mispredictions on either class, and examples falling in between the hyperplanes can be naturally rejected~\citep{Varshney2006}.
With the same approach, one can also learn a \gls{ocsvm} for each class to reject test examples that lie outside any learned hypersphere (novelty) or within two overlapping hyperspheres (ambiguity)~\citep{Lotte2008,Loeffel2015,Wu2007}. Finally, \citet{Sousa2014a} shows that limiting the number of support vectors reduces the computational cost, while still ensuring high performance in most cases.

However, in several cases, more than two SVMs are used. For instance, in multi-label classification one can exploit as many SVMs as the number of labels and fit each hyperplane to discriminate between one class and all the others~\citep{Pillai2011}. This raises the issue of defining rejection in the regions of intersection between some, but not all, of the hyperplanes. To address this, a natural solution is to utilize a data-replication method~\citep{Sousa2009b}. This approach involves replicating the complete dataset for each class $k \in \mathcal{Y}$, adding a new dimension $z$ with the class number, changing the target variable of each replica to a discrete one-vs-all label, and discriminating class $k$ from the other classes~\citep{Cardoso2007,DaRochaNeto2011}.

Finally, \emph{Neural Network models}  allow integrating the rejector and predictor into the same structure by modifying their output layers~\citep{Gangrade2021,Ziyin2019}.  \citet{Geifman2019b} propose to introduce an additional head $m_r$ into the network that is dedicated to rejection. Specifically, $m_r$ is set as a sigmoid function and used such that 
\begin{equation*}
    m(x) = \rr{} \quad \text{ if } m_r(x) < 0.5.
\end{equation*}
Similar to the SVM case, \citet{GascaA.2011} and \citet{Mesquita2016} measure the disagreement of two Neural Networks trained to prioritize the classes differently. Specifically, they assume a binary classification task and use the output of two neural networks $h_1$, $h_2$ to predict the positive class if $h_1(x), h_2(x) \ge 0$, the negative class if $h_1(x), h_2(x) < 0$ and rejection if they disagree on the sign. For this task, they use two weighted Extreme Learning Machines (wELM)~\citep{Zong2013}, namely two neural networks with $Q$ hidden neurons that output 
$$h_*(x) = \sum_{q=1}^Q w_y \beta_q g(a_q \cdot x + b_q)$$
where $w_y$ is the cost related to the example $x$ that belongs to the class $y$, $a_q$ is the weight vector connecting the $q$-th hidden node and the input nodes, $b_q$ is the bias of the $q$-th hidden node and $g$ is the activation function. By setting the class misprediction costs, learning the parameters $\beta = (\beta_1,\dots,\beta_Q)$ requires using the traditional weighted least square formulation $\min \| H \beta - Y\|^2$ so that
\begin{equation*}
    \beta = (H^TWH)^{-1}H^TWY
\end{equation*}
where $H$ is the $n\times Q$ matrix of activation functions $h_{iq} = g(a_q \cdot x_i + b_q)$, $W$ is the $n\times 1$ matrix of class costs (one per example) and $Y$ is the target vector. Thus, each network limits one class mispredictions and the region of disagreement is designed to be the rejection region.

\paragraph{\bf{Benefits and drawbacks of an integrated rejector.}}
This architecture has two key \emph{benefits}.
First, integrating the predictor and rejector means that both aspects of the model are optimized toward the task at hand. 
This can improve the performance of the model with rejection when compared to using other architectures because the predictor's and the rejector's components can affect each other.
Second, because it is a unique model, the bias introduced by the model with rejection is potentially less than in the scenario where the predictor and rejector are two different models. 
However, this architecture has potential \emph{drawbacks}, as designing such a rejector might not be trivial. First, it requires extensive knowledge about the predictor in order to integrate the reject option. Second, it may require developing a novel algorithm to learn the model with a reject option from data. Finally, it is computationally more expensive than the other architectures, as any changes to the rejector require retraining the entire model, which can be time-consuming~\citep{Clertant2020,Shpakova2021}.

%% file: g_CombineRejectors.tex
\section{Combining multiple rejectors}\label{sec:combining_rejectors}

Most rejectors are tailored towards a single rejection type. However, by combining multiple rejectors one can enable multiple rejections, such as performing both ambiguity and novelty rejection. We distinguish between two types of combinations of rejectors based on whether the rejectors' rejection regions overlap or not because examples in overlapping regions require deeper analysis (e.g., to specify the underlying rejection type). 

First, when rejectors do not overlap (or when we are not interested in the example's rejection type), one can simply combine the rejection sets by a logical $or$-rule: reject the example if any of the rejectors rejects it and assign such rejection type~\citep{Frelicot1997,Suutala2004}. For instance, given $\mathcal{Z}$ rejectors $r_1, r_2, \dots, r_{\mathcal{Z}}$ with thresholds $\tau_1, \tau_2, \dots, \tau_{\mathcal{Z}}$, one can combine them into $m$ as
\begin{equation*}
    m(x) =
    \begin{cases}
        \rr{} \quad &\text{if } \exists \, i \le \mathcal{Z} \, \colon r_i(x; h)<\tau_i \\
        h(x)  \quad &\text{otherwise}
    \end{cases}
\end{equation*}

\changed{Second, when rejectors overlap in some regions a simple $or$-rule can be insufficient to determine the reason for rejection because each rejector may decide to abstain for a different reason.} Typically, existing works carefully select the order to evaluate the rejectors. This is usually done in a multi-step architecture: either by stacking only the rejectors~\citep{Frelicot1998,Frelicot2002}, or even using multiple models with rejection~\citep{Pudil1992,Barandas2022}. For instance, given $\mathcal{Z}$ rejectors $r_1, r_2, \dots, r_{\mathcal{Z}}$ with thresholds $\tau_1, \tau_2, \dots, \tau_{\mathcal{Z}}$, one can order rejectors by importance and combine them as
\begin{equation*}
    m(x) =
    \begin{cases}
        \rr{}_1 \quad &\text{if } r_1(x; h)<\tau_1; \\
        \rr{}_2 \quad &\text{if } r_1(x; h)\ge \tau_1 \text{ and } r_2(x; h)<\tau_2; \\
        \vdots \\
        h(x)  \quad &\text{if } r_i(x; h)\ge \tau_i \ \forall \, i \le \mathcal{Z}; \\
    \end{cases}
\end{equation*}
where $\rr{}_i$ indicates the rejector $r_i$'s type of rejection.

%% file: h_applications.tex
\section{Applications of machine learning models with rejection}~\label{sec:applications}
In safety-sensitive domains, making the wrong decision can have serious consequences such as fatal accidents with self-driving cars, major breakdowns in industrial settings or incorrect diagnoses in medical applications. In these domains, rejection can be used to make cautious predictions. However, the number of papers discussing machine learning with rejection in practical applications is still limited. In this section, an overview of these papers is given. 

\paragraph{Biomedical applications.} Machine learning with rejection is primarily explored in medical applications due to the potential consequences of incorrect decisions~\citep{Liu2022}. The main focus is on ambiguity rejection for medical diagnosing, specifically the detection and classification of diseases. If the model is confident enough, the detection results are automatically translated into a diagnosis. Otherwise, a medical expert verifies the detection~\citep{Kompa2021}. For instance, vocal pathologies are detected using voice recording data, where a linear classifier is trained and uncertain predictions are rejected based on a threshold of the derived posterior probability~\citep{Kotropoulos2009}. Spine disease diagnosis employs the data-replication method, which predicts only when two biased classifiers agree~\citep{Sousa2009b}. Cancer detection, particularly breast tumor detection, benefits from a reject option implemented with an SVM classifier using confidence-based rejection~\citep{Guan2020}. The rejection thresholds are chosen to limit the rejection rate and reduce manual effort. 

Other biomedical applications have also adopted the use of a reject option. \citet{Lotte2008} conduct an experiment on distinguishing hand movements using brain activity. They employ a separate novelty rejector trained in a supervised manner to discard brain activity associated with other activities. \citet{Lewicke2008} explore sleep stage scoring with both types of rejection, utilizing confidence metrics derived from a Neural Network classifier's neural activities. Another sleep stage scoring application utilizes a separate rejector based on \gls{lof} anomaly scores for novelty rejection, identifying patients who deviate from the training data~\citep{VanderPlas2021}. Some papers compare the performance of multiple models with rejection to determine the optimal approach for specific biomedical applications. For instance, \citet{Kang2017} predict the effectiveness of a diabetes drug for individual patients, while \citet{Tang2014} investigate the classification of body positions using sensors placed in patients' shoes. Lastly, a medical application focuses on the analysis of tissue examples, aiming to classify each pixel of tissue images into categories such as bone, fat, or muscle~\citep{Condessa2013,Condessa2015a}.

\paragraph{Engineering applications.} 
Applications in engineering can also benefit from a reject option. For instance, in the chemical identification of gases, time-series data is processed by two classifiers to classify the observed gas. Classification occurs only when there is agreement between the classifiers, and ambiguous predictions are rejected until consensus is reached~\citep{Hatami2013}. A similar ambiguity rejection technique, rejecting when two classifiers disagree, is employed in defect detection in software applications~\citep{Mesquita2016}. Fault detection in steam generators utilizes a set of one against all SVM classifiers, and rejection is based on the distance to the decision boundary of these classifiers, allowing for both rejection types~\citep{Zou2011}. Finally, \cite{Hendrickx2022} employs a separated novelty rejector for vehicle usage profiling.

\paragraph{Economics applications.} In the domain of economics, two applications of machine learning with rejection have been proposed, both focused on ambiguity rejection and novelty rejection. The first application uses a \gls{lvq} to classify dollar bills by value~\citep{Ahmadi2004}. Confidence metrics are obtained from the classifier for both types of rejection. The second application investigates a few rejection techniques on top of a predictor to decide whether to grant a loan~\citep{Coenen2020}

\paragraph{Image recognition applications.} Reject options are usually available for analyzing text styles and reading handwritten numbers~\citep{Fumera2004}, used for both ambiguity rejection~\citep{Xu1992,Huang1995,RAHMAN1998} and novelty rejection~\citep{Lou1999c,Arlandis2002}. These methods employ a confidence-based dependent rejector. Additionally, there is a paper focused on identifying walkers based on their footprints~\citep{Suutala2004}. Initially, each footprint is individually predicted or rejected, and then the information from three consecutive footprints is combined for the final decision.

%% file: i_related_areas.tex
\section{Link to other research areas}\label{sec:related_research_areas}

This section briefly discusses the fields related to learning with rejection. 
\subsection{Uncertainty quantification}
    \label{sec:uncertainty}

The field of uncertainty quantification \changed{(UQ)} aims to measure how uncertain \changed{a learned model's predictions are}~\citep{Kendall2017}. It distinguishes between two types of uncertainties: aleatoric uncertainty, which is the randomness in the data, and epistemic uncertainty, which is the lack of knowledge. Aleatoric uncertainty arises from non-deterministic relations between features and the target, while epistemic uncertainty can be caused by a small training set or incorrect model bias. For instance, when predicting the outcome of tossing an unfair coin, initially we lack historical data, resulting in high data-epistemic uncertainty. As we observe more coin tosses, data-epistemic uncertainty decreases, but aleatoric uncertainty remains due to the stochastic nature of the coin flip~\citep{Senge2014}.

\changed{These uncertainties are inherently related to rejection. Rejecting examples due to high aleatoric uncertainty falls into the ambiguity rejection scenario. On the other hand, high epistemic uncertainty due to the lack of data may cause either ambiguity or novelty rejection. That is, if an example is similar to the training set but its prediction strongly depends on the choice of the dataset (e.g., close to the predictor's decision boundary for classification tasks), then this gives rise to an ambiguity rejection.
Alternatively, if an example is dissimilar to any of the training examples, this leads to a novelty rejection.
}

\changed{Methods for UQ can be applied within learning with rejection.
UQ focuses on obtaining (calibrated) estimates that meaningfully convey the level of uncertainty~\citep{Kotelevskii2022}, which learning with rejection can leverage to allow the model to abstain when the uncertainty is high~\citep{Perello-Nieto2017,Kompa2021}. While calibrated uncertainty estimates are not always necessary for learning with rejection, they can be important. For instance, calibrated uncertainty estimates enable setting an optimal threshold that minimizes the empirical risks~\citep{Chow1970a}.
}


\subsection{Anomaly detection}

Anomaly detection~\citep{Prasad2009} is a Data Mining task aimed at identifying examples that deviate from expected behavior in a dataset. It is closely linked to novelty rejection because anomalies, being rare and substantially different from the training data, fall under the category of novelties~\citep{Ulmer2020,Pimentel2014,Markou2003}. Anomaly detectors are often utilized for novelty rejection within a separate rejector architecture.

Adding a reject option to anomaly detectors allows them to abstain from processing examples when a clear decision cannot be made~\citep{Perini2020,Perini2023}. However, enabling this option in unsupervised anomaly detection poses two challenges. First, most confidence metrics assume a supervised setting, relying on measuring the distance to a decision surface. However, in anomaly detection, a hard decision surface may not always exist, necessitating specialized metrics that consider the model bias of the detector~\citep{Perini2020}. Second, the lack of labeled data makes it difficult to train a rejector using standard performance metrics. Instead, unsupervised techniques are employed, leveraging performance metrics that measure the stability of the anomaly detector itself~\citep{Perini2020a}.

\subsection{Active learning}

Active learning~\citep{Settles2009b,Fu2013,Zhang2014,Nguyen2019b} involves the interaction between a learning algorithm and an oracle who provides feedback to guide the learner. Its purpose is to reduce the need for labeling large amounts of data while still achieving high predictive performance. The algorithms focus on identifying the examples that would be most beneficial for the learner to label, thus minimizing the associated labeling costs.

Active learning and learning with rejection share the focus on uncertain examples~\citep{Amin2021}. However, they differ in their motivations for addressing uncertainty. In active learning, uncertainty is crucial during training to improve efficiency by minimizing the amount of labeled data required for an accurate model. In contrast, learning with rejection aims to capture uncertainty at test time to prevent mispredictions. Its focus is on avoiding unreliable predictions based on uncertain examples. 

\changed{Another difference is that outliers are not always considered. For instance, methods based on discriminative learning cannot express low-density regions. \citet{Sharma2017} determine uncertain examples based on evidence measures that support the positive ($E_{+1}$) or negative class ($E_{-1}$) in binary classification. An example $x$ has an uncertain class if $E_{+1}(x) \approx E_{-1}(x)$. Two cases are distinguished based on the magnitude of the evidence: if both $E_{+1}$ and $E_{-1}$ are large, the model is uncertain because of strong, but conflicting evidence for both classes, while if both $E_{+1}$ and $E_{-1}$ are small, the model is uncertain because of insufficient evidence for either class. Because both cases assume that a model is uncertain if $P(Y|X)\approx 0.5$ when using a uniform prior, they both correspond to our ambiguity rejection scenario.
%
%
}

\changed{Combining active learning with machine learning with rejection} could be of great use~\citep{Korycki2019,Puchkin2022,Shekhar2020a}. When the interaction with an oracle is possible, it may be of interest to query the rejected test examples. New data types could be identified by novelty rejection, while ambiguity rejection may fine-tune the decision boundary.

\subsection{Class-incremental / incomplete learning}

Typically, learned models assume knowledge of all possible classes during training. However, class-incremental learning focuses on models that adapt during deployment to detect and predict novel classes that were not seen during training.

Novelty rejection and class-incremental learning both operate under an open-world assumption and aim to detect novel examples compared to the training set. However, there are two key differences. First, class-incremental learning specifically targets examples belonging to novel classes, distinguishing them from outliers. In contrast, novelty rejection techniques do not prioritize this distinction. Second, class-incremental learning involves detecting novel class examples and retraining the model to recognize them. 

Novelty rejection techniques can be considered in class-incremental learning. Moreover, both techniques can be combined into a single pipeline, by adapting incremental models with novelty rejected examples. For instance, such examples can be used as prototypes in a \gls{knn} model.

\subsection{Delegating classifiers}

\changed{Similar to learning with rejection, the approach of delegation involves the use of a classifier, which only classifies examples with high confidence and delegates the prediction for the remaining examples~\citep{Temanni2007,Khodra2016}. The delegated examples are given to another, more specialized, classifier which makes a prediction~\citep{Ferri2004a,Prasad2008}. In contrast, learning with rejection usually assumes that the user will inspect any rejected examples. Furthermore, delegation can be developed as a chain, where the next classifier makes a prediction for the examples for which the previous model was too uncertain~\citep{Giraud-Carrier2022}.}

\subsection{Meta-learning}
Meta-learning, also known as ``learning to learn'', explores methods and techniques for automatically learning the characteristics, behaviors, and performance of machine learning models~\citep{Bock1988,Vanschoren2018}. It aims to develop higher-level knowledge that guides the learning process itself~\citep{Brazdil2009,Gridin2022}.

Despite having different goals and levels of abstraction, meta-learning can provide valuable insights and approaches for the context of learning \changed{with rejection}. For instance, meta-learning algorithms analyze the behavior and performance of classifiers on different datasets to derive general knowledge about their strengths, weaknesses, and limitations~\citep{Abbasi2012,Tremmel2022,Cohen2022}. This knowledge can then be used to make informed decisions about when to reject predictions. Moreover, meta-learning algorithms can identify relevant features or attributes that are informative for determining when to reject predictions~\citep{Filchenkov2016,Shen2020a}. By focusing on important features, rejectors can make more accurate decisions.

%% file: j_conclusions.tex
\section{Conclusions and perspectives}\label{sec:conclusion_and_future_direction}

We have studied the subfield of machine learning with rejection and provided a higher-level overview of existing research. To conclude, we revisit our key research questions and point to new directions that future research might take.

\subsection{Research questions revisited}
This survey paper is built around eight key research questions, introduced in the introduction. In this section, we revisit each of these questions and briefly summarize our findings.

\paragraph{How can we formalize the conditions for which a model should abstain from making a prediction?}
In Section~\ref{sec:preliminaries}, we identify two types of rejections: ambiguity rejection and novelty rejection.
Ambiguity rejection abstains from making a prediction an example falls in a region where the target value is ambiguous (e.g., close to the decision boundary in classification tasks).
This could be due to a non-deterministic true relation between the features and target variable, or due to a hypothesis space that is not able to capture the true relation. 
Novelty rejection abstains from making a prediction on examples that are rare with respect to the given training set. For such an example, there is no guarantee that the model correctly extrapolates to this untrained region, making it likely that the model mispredicts the example.

\paragraph{How can we evaluate the performance of a model with rejection?}
Standard machine learning evaluation is focused on a model's predictive quality. However, in machine learning with rejection, there exists a trade-off between the predictive quality and the proportion of rejected examples.

In Section~\ref{sec:metrics}, we provide an overview of techniques evaluating both the prediction and rejection quality of models with rejection. We identify three categories: metrics evaluating models with a given rejection rate, metrics evaluating the overall model performance/rejection trade-off, and metrics evaluating models through a cost function.

\paragraph{What architectures are possible for operationalizing (i.e., putting this into practice) the ability to abstain from making a prediction?}
We categorize machine learning methodologies with rejection in three different architectures, depending on the relationship between the predictor and the rejector: separated, dependent and integrated rejector. These categories are introduced and mapped to the existing literature in Sections~\ref{sec:separated_rejector},~\ref{sec:dependent_rejector} and~\ref{sec:integrated_rejector}. 

\paragraph{How do we learn models with rejection?}
For each architecture, we discuss the main techniques to learn a model with rejection and related these to the existing literature in Sections~\ref{sec:separated_rejector},~\ref{sec:dependent_rejector} and~\ref{sec:integrated_rejector}. 
First, the separated rejector is usually learned independently of the predictor.
Second, learning the dependent rejector entails learning for which examples the predictor is likely to mispredict using a confidence function. Both architectures need setting a rejection threshold.
Third, integrated rejector needs a unique algorithm for learning predictor and rejector in tandem. Usually, this architecture relies on designing an objective function.

\paragraph{What are the main pros and cons of using a specific architecture?}
Each architecture, discussed in Sections~\ref{sec:separated_rejector},\ref{sec:dependent_rejector}, and~\ref{sec:integrated_rejector}, offers distinct benefits and drawbacks. Separated rejectors show broad applicability, as they can be combined with any predictor. However, they often yield sub-optimal rejection performance since they do not learn from the predictor's mispredictions. On the other hand, dependent rejectors have reduced, yet still high, applicability, relying on a specific confidence function learned from the predictor's output, but they can enhance the rejection quality by leveraging the predictor's mispredictions. Finally, integrated rejectors necessitate joint design with the predictor, but learning a single model for prediction and rejection improves the overall performance for both prediction and rejection tasks.

\paragraph{How can we combine multiple rejectors?}
We discuss the combination of multiple rejectors for enabling various types of rejections within a unique model. There are two approaches for combining rejectors. First, when rejectors do not overlap, a logical ``or'' rule is applied, rejecting an example if any of the rejectors rejects it. Second, when rejectors overlap in some regions and disagree on the type of rejection, a multi-step architecture is used, ordering rejectors by importance to make decisions based on the most relevant rejector.

\paragraph{Where does the need for machine learning with rejection methods arise in real-world applications?}
On a high level, machine learning with rejection is typically used in applications where incorrect decisions can have severe consequences, both financially and safety-related. These consequences motivate the need for robust and trustworthy machine learning. In Section~\ref{sec:applications}, we provide an overview of application areas in which machine learning with rejection is already used.

\paragraph{How does machine learning with rejection relate to other research areas?}
Section~\ref{sec:related_research_areas} shows that machine learning with rejection is closely related to several other subfields of machine learning. This relation sometimes leads to terminology and techniques overlapping or inspired by these other domains. In contrast, other cases show machine learning with rejection from a broader perspective. In this survey, we related machine learning with rejection to uncertainty quantification, anomaly detection, active learning, class-incremental learning, delegating classifiers, and meta-learning.

\subsection{Future directions}
Given its significance for the usage of machine learning in real-world problems and the growing attention for trustworthy AI, we expect machine learning with rejection to remain an active research field. In this section, we briefly discuss three key research directions for which we see a strong need.

\paragraph{Standard settings to compare different models with rejection.}
A large number of machine learning models with rejection already exist. However, these are typically evaluated on custom or even proprietary data. This makes it difficult to benchmark and compare the different approaches. While some papers use publicly available datasets, there is no standard benchmark set for machine learning with rejection.
Additionally, applying multiple strategies to evaluate the rejector offers a better view of an algorithm's performance and improves comparability.

\paragraph{Partial rejection for machine learning models.}
A promising avenue for further exploration is the concept of partial abstention. 
\changed{Nowadays, machine learning problems often involve seeking elaborated predictions rather than simple scalar or class values as in classification and regression tasks. For instance, in multi-label classification, a prediction for an instance is a subset of possible class labels.
In such cases, the idea of abstaining from a complete prediction can be extended to partial abstention, where the learner delivers predictions on \emph{some but not necessarily all}
class labels, according to its level of certainty~\citep{Nguyen2020}. 
}
This has the key benefit of providing a middle ground between making predictions for the entire structure and completely abstaining from making any predictions.

\paragraph{Algorithms enabling models with rejection in domains other than classification.}
Most papers on machine learning with rejection study supervised classification problems. Modern machine learning tackles numerous other tasks such as regression, forecasting, and clustering, or even semi-supervised and self-supervised feedback loops. We believe that the rejection can also be of use in such areas. However, since only a handful of relevant studies exist, this requires more attention from the research community. 
Future research can also focus on integrating rejection-related variables into statistical frameworks utilized in educational measurement, such as Item Response Theory (IRT). This integration has the potential to improve the precision of item calibration, trait estimation, and the interpretation of test scores. Furthermore, it can provide valuable insights into the psychological aspects of learning, enabling the development of more precise instructional strategies and interventions.